\documentclass[letterpaper, 10 pt, conference]{ieeeconf}  

\IEEEoverridecommandlockouts                              

 \usepackage{amsmath} 

\usepackage{amsmath} 
\usepackage{amssymb}  
\usepackage{booktabs}
\usepackage[table]{xcolor} 
\usepackage{multirow}
\usepackage{siunitx}
\usepackage{xcolor}
\usepackage{tabularx}
\usepackage{graphicx}
\usepackage{makecell}
\usepackage{pifont}

\title{\LARGE \bf
MTRDrive: Memory-Tool Synergistic Reasoning for Robust Autonomous Driving in Corner Cases
}

\author{
    Ziang Luo$^{1,2*}$, 
    Kangan Qian$^{1,2*}$, 
    Jiahua Wang$^{2,\dagger}$, 
    Yuechen Luo$^{1}$, 
    Jinyu Miao$^{1}$, 
    Zheng Fu$^{1}$, \\
    Yunlong Wang$^{1}$,
    Sicong Jiang$^{3}$,
    Zilin Huang$^{4}$,
    Yifei Hu$^{2}$,
    Yuhao Yang$^{2}$,
    Hao Ye$^{2}$,
    Mengmeng Yang$^{1}$, \\
    Xiaojian Dong$^{2}$,
    Kun Jiang$^{1}$,
    Diange Yang$^{1,\ddagger}$%
\thanks{$^{1}$School of Vehicle and Mobility, Tsinghua University, Beijing, China}
\thanks{$^{2}$Automotive and Robotics, Xiaomi Corporation, Beijing, China}
\thanks{$^{3}$McGill University, Montreal, Quebec, Canada}
\thanks{$^{4}$University of Wisconsin-Madison, Madison, WI, USA}
\thanks{$^{*}$Equal contribution. $^{\dagger}$Project Leader. $^{\ddagger}$Corresponding authors}}

\begin{document}
\maketitle
\thispagestyle{empty}
\pagestyle{empty}

\begin{abstract}

Vision-Language Models(VLMs) have demonstrated significant potential for end-to-end autonomous driving, yet a substantial gap remains between their current capabilities and the reliability necessary for real-world deployment. A critical challenge is their fragility, characterized by hallucinations and poor generalization in out-of-distribution (OOD) scenarios.
To bridge this gap, we introduce MTRDrive, a novel framework that integrates procedural driving experiences with a dynamic toolkit to enhance generalization and proactive decision-making.

MTRDrive addresses these limitations through a closed-loop system that combines a memory-based experience retrieval mechanism with dynamic toolkits. This synergy enables the model to interact more effectively with its environment, improving both reasoning and decision-making capabilities with the help of our memory-tool synergistic reasoning. Additionally, we introduce a new benchmark based on complex Roadwork construction scenarios to rigorously evaluate zero-shot generalization.

Extensive experiments demonstrate the superior effectiveness of our approach. On the public NAVSIM benchmark, our 3B-parameter MTRDrive model achieves an exceptional PDMS of 88.3 without chain-of-thought and sets a state-of-the-art performance bar on high-level planning, with a driving metric score of 79.8\% and a planning accuracy of 82.6\%. Rigorous zero-shot evaluation on the new Roadwork-VLM benchmark shows a strong ability to reason robustly in unseen scenarios, achieving a driving metric score of 80.2\%. These results highlight MTRDrive's potential to advance autonomous driving toward safer and more reliable systems.
\end{abstract}


\section{INTRODUCTION}

\begin{figure}[ht!]
\centering
\includegraphics[width=0.5\textwidth]{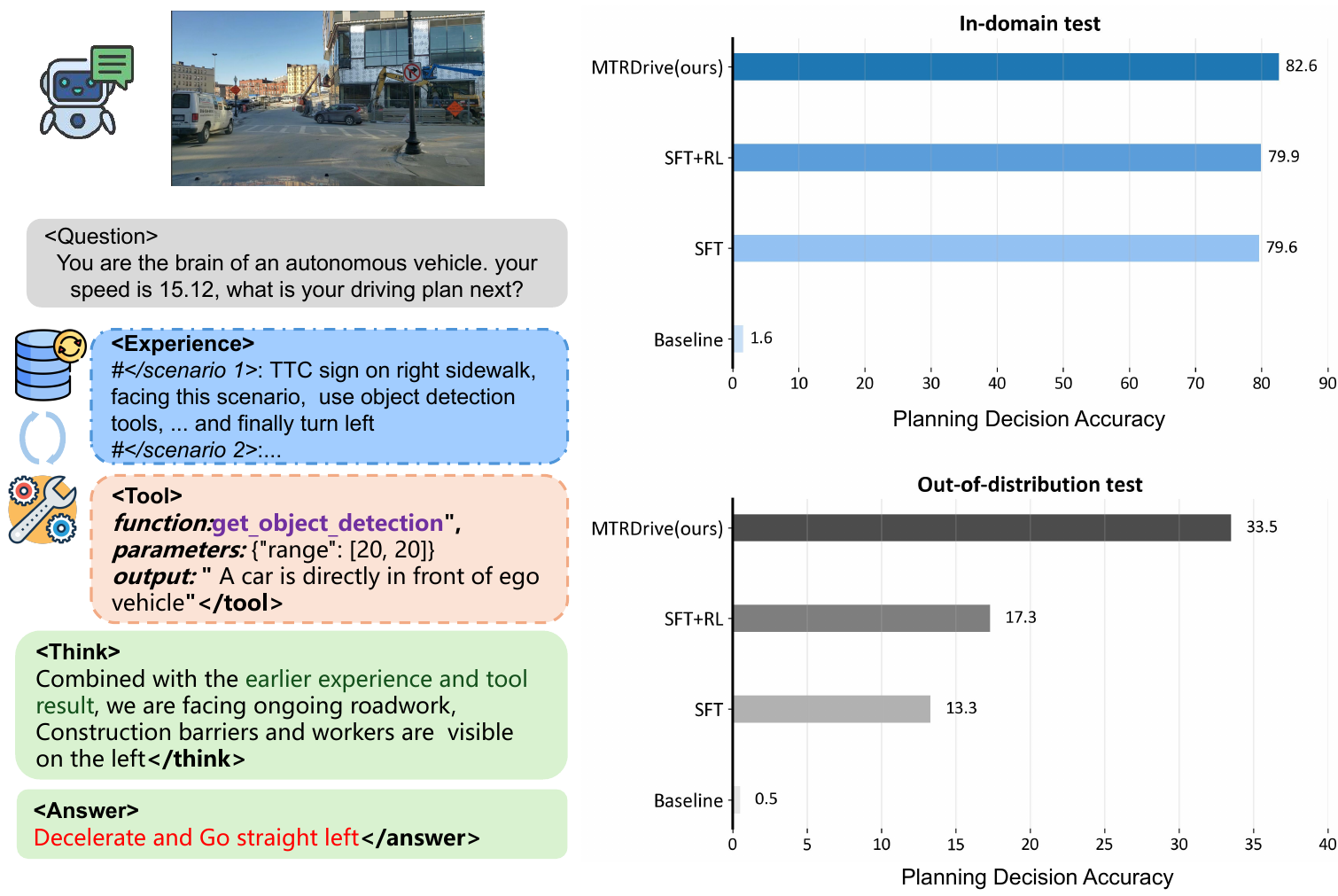}
\caption{
An overview of our MTRDrive framework and its performance benefits.
\textbf{Left:} The interactive reasoning process, where the agent leverages retrieved past experiences and real-time tool use in a chain-of-thought process to make a final driving decision.
\textbf{Right:} Quantitative results show that MTRDrive significantly boosts planning decision accuracy over baseline methods. The improvement is particularly dramatic in the challenging out-of-domain test, demonstrating enhanced generalization.
}
\label{fig:intro_overview}
\end{figure}
The emergence of Vision-Language Models (VLMs)~\cite{jiang2025vlasurvey} has propelled the paradigm of end-to-end autonomous driving forward, building a single, cohesive system that mirrors human cognitive processes. Unlike traditional modular pipelines that separate perception\cite{shi2024effocc, qian2024priormotion}, prediction\cite{qian2025lego}, and decision-making\cite{qian2024spider, fu2025enhancing}, VLMs break down these divisions by jointly modeling multi-modal inputs and generating holistic driving responses in a unified framework. 
Recent impressive demonstrations show that these models, often enhanced by techniques like Chain-of-Thought (CoT), can achieve strong performance across various tasks, from perception-based Visual Question Answering (VQA) \cite{qian2025agentthink, nie2024reason2drive, shao2024lmdrive, sima2024drivelm} to intricate motion planning \cite{li2025recogdrive, fu2025orion, zhou2025autovla}.

Despite these advances, a significant gap remains between current VLM performance and the reliability required for real-world deployment: The models are inherently fragile, often exhibiting visual hallucinations and failing in out-of-distribution (OOD) scenarios \cite{qian2025agentthink, ghosh2024roadwork}. In autonomous driving, a field where even a minor error can lead to catastrophic outcomes, such limitations present a critical barrier to adoption\cite{qian2024fasionad, qian2025fasionad++}.

Fundamentally, a robust driving decision is influenced by two key factors: the accuracy of perception and the soundness of reasoning. Human cognition itself can be viewed as a lifelong feedback loop:
\begin{quote}
    \textit{We are perpetually in a state of self-query: Our initial experiences form the 'first draft' of this model, which is then continuously edited and refined by every subsequent interaction throughout our lives.}
\end{quote}
This principle is directly embodied by a human driver, who constantly engages in a closed-loop interaction with their environment. They ensure safety through continuous perception and combine real-time analysis and past driving experience to judge and execute their next action. This dynamic, interactive process of observing and reflecting is what enables reliable navigation in complex and unforeseen situations.

Inspired by this human-like process, we propose MTRDrive, a novel framework built on the principle of Interactive Reasoning to address the limitations of current VLM-based driving agents. Unlike traditional methods that treat each input as a one-shot decision, MTRDrive endows the agent with the ability to proactively retrieve driveing experience and employ tools to query its environment. As vividly demonstrated in Figure \ref{fig:intro_overview}, by shifting from a static decision-making model to a dynamic, interactive one, our approach mitigates the risk of hallucinations and dramatically improves performance in novel, unseen scenarios.


In summary, our key contributions are:

\begin{itemize}
    \item We introduce \textbf{MTRDrive}, a novel framework that models autonomous driving as a dynamic, interactive process, moving beyond static, one-shot decision-making paradigms.
    \item We propose a \textbf{memory-tool synergy} mechanism that enables the agent to utilize both a knowledge base of past driving experiences and active toolkits for real-time information retrieval.
    \item We demonstrate that our approach effectively \textbf{mitigates visual hallucinations} and \textbf{improves generalization} in challenging out-of-distribution scenarios, which is critical for real-world deployment.
\end{itemize}

\section{Related Work}
\subsection{Vision-Language-Action Models for Autonomous Driving}
Recent advancements in language modeling have created new opportunities for autonomous driving, particularly through vision-language-action (VLA) systems. Early efforts integrated Visual Language Models (VLMs) with textual prompts to generate scene descriptions and high-level navigation instructions \cite{mao2023gptdriver, jiang2024senna} without directly producing control signals. For example, DriveGPT-4 \cite{xu2024drivegpt4} processed a single front-camera image to produce either textual descriptions or high-level maneuver labels (e.g., "slow down," "turn left"), while actual vehicle control remained handled by traditional modules such as PID controllers. Subsequent work, including OpenDriveVLA \cite{zhou2025opendrivevla} and Orion \cite{fu2025orion}, significantly narrowed the semantic gap between language instructions and vehicle actions, effectively embedding natural language into the core planning loop. To further advance end-to-end pipeline performance, unified VLA models such as EMMA \cite{hwang2024emma}, LMDrive \cite{shao2024lmdrive}, CarLLaVA \cite{renz2024carllava}, and SimLingo \cite{renz2025simlingo} have been developed based on language backbones and fine-tuned using trajectories from human experts or simulators.

However, these data-driven fine-tuning methods often struggle to mitigate visual hallucinations and perform inadequately in out-of-distribution (OOD) scenarios. In contrast, our framework empowers the agent to proactively interact with its environment with memory-tool synergistic reasoning.

\subsection{Chain-of-Thought in Autonomous Driving}
The latest generation of VLMs extends beyond explanation and plan conditioning to support long-horizon reasoning. Chain-of-Thought (CoT) methodologies in autonomous driving have emerged as an explainable and transparent paradigm to improve perception, cognition, and planning capabilities. Studies such as DriveCoT \cite{wang2024drivecot}, AgentDriver \cite{mao2023agentdriver}, and Sce2DriveX \cite{zhao2025sce2drivex} employ step-by-step reasoning with fixed prompt templates, enriching contextual information and leading to improved driving decisions in complex scenarios. AgentThink \cite{qian2025agentthink} introduced a framework that integrates tool-use capabilities into the VLM reasoning process, allowing models to dynamically invoke tools to enhance reasoning accuracy and reduce hallucinations.
AutoVLA \cite{zhou2025autovla} and Impromptu VLA\cite{chi2025impromptu} fuse reasoning processes with trajectory planning within the CoT framework, demonstrating superior generalization across benchmarks such as NAVSIM \cite{dauner2024navsim} and Bench2Drive \cite{jia2024bench2drive}.

However, these methods often depend on extensive CoT data to develop reasoning capabilities, which is a process that is highly inefficient and prone to hallucinations during reasoning. Moreover, they still fall short in leveraging structured knowledge inherent to the autonomous driving domain\cite{wang2023drivemlm}. Such limitations impede the ability of self-driving vehicles to perform robust detection and planning in complex environments. By contrast, our approach introduces a driving knowledge base for efficient experience reuse and a tool-calling agent that supplies quantitative outputs (e.g. detection results), ultimately supporting a more interactive and grounded reasoning process.

\section{Methodology}
\begin{figure*}[h!]
    \centering
    \includegraphics[width=0.95\linewidth]{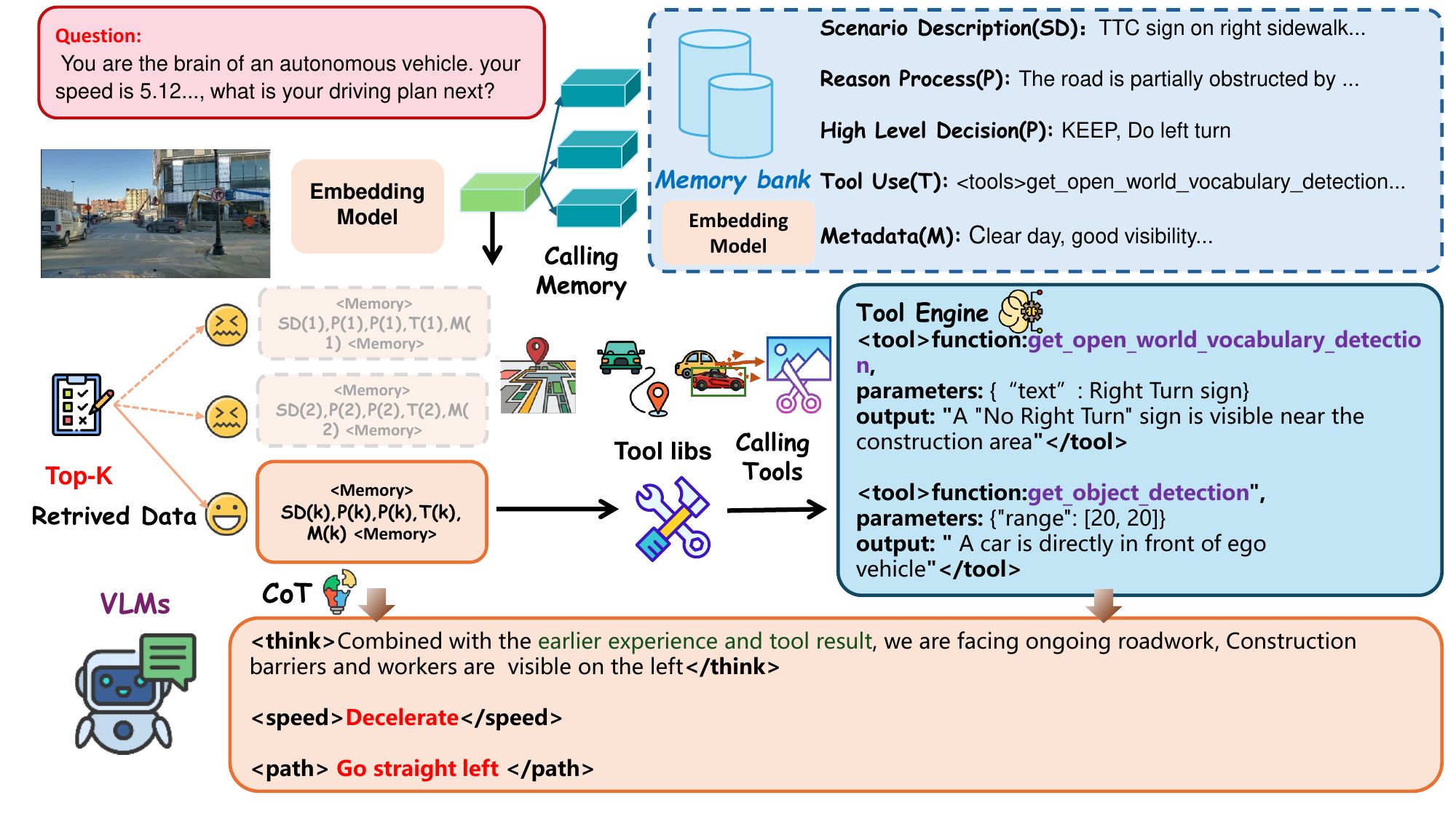}
    \caption{Overview of MTRDrive. Given a forward-facing camera image and a text prompt, the Vision-Language Model (VLM) interacts with two external modules: a memory bank for experience retrieval and a tool engine for information gathering. The retrieved knowledge and tool-generated results are synthesized within a chain-of-thought reasoning process to produce the final driving actions, such as speed and path adjustments.}
    \label{fig:architecture}
\end{figure*}

Our proposed framework, MTRDrive, is a Vision-Language Model (VLM) specifically designed for autonomous driving planning. Drawing inspiration from human cognitive processes, MTRDrive enhances its decision-making capabilities by integrating memory retrieval and tool interaction. By retrieving and reasoning over past driving experiences, the model improves its planning accuracy and grounds its tool selection process in relevant prior knowledge, thereby mitigating the risk of visual hallucinations. The overall architecture of our framework is depicted in Figure \ref{fig:architecture}.

The MTRDrive framework comprises two core components: a Driving Experience Base that stores structured driving experience, and an Experience-Driven Planning Module that leverages these experience for tool interaction and decision-making. This architecture enables our agent to achieve a higher level of cognitive function by synergizing historical knowledge with real-time perception.


\subsection{Driving Experience Base Construction}
To emulate the human ability to learn from experience, we first construct a structured, retrievable knowledge base of past driving scenarios.
\subsubsection{Semantic Scenario Encoding}
The foundation of our experience base is an effective retrieval system that can identify semantically similar past scenarios. While generative VLMs like Qwen-VL\cite{bai2025qwen25} excel at multimodal understanding, their complex architectures and input formats are suboptimal for the large-scale, low-latency similarity retrieval required for real-time driving. Therefore, we require an embedding model that prioritizes semantic representation efficiency.

We selected the pre-trained visual encoder from CLIP\cite{radford2021clip} for this purpose. CLIP's architecture is explicitly optimized for learning a joint image-text embedding space, making it highly effective at capturing visual semantic similarity. Compared to large generative models, the CLIP encoder offers significantly faster inference speeds and a smaller memory footprint, rendering it ideal for our large-scale retrieval task. Its training on a vast dataset of image-text pairs ensures a rich and generalizable feature space without requiring task-specific fine-tuning.

Formally, for any given driving scenario image $I$, the visual encoder $\mathcal{F}_{enc}$ maps it to a latent embedding vector $v_I = \mathcal{F}_{enc}(I)$, where $v_I \in \mathbb{R}^d$. The similarity between two scenarios, represented by images $I_1$ and $I_2$, is computed using the cosine similarity of their embeddings:

$$\text{Similarity}(I_1, I_2) = \frac{v_{I_1} \cdot v_{I_2}}{\|v_{I_1}\| \|v_{I_2}\|}$$

This efficient and semantically meaningful representation serves as the key for retrieving relevant experiences from our knowledge base.

\subsubsection{Structured Experience Representation}
The structured knowledge base formally represents each experience as a Driving Scenario Document. This document, which we use as a tuple for conceptual clarity, stores the key elements of past driving experience as $\langle SD, P, H, T, M \rangle$. Here, $SD$ is the scenario description, which defines the scanarios encountered by the agent; $P$ is the reasoning process, an abstract solution trajectory that includes the agent's thought process and the invocation of tools; $H$ is the high-level decision ($\gamma$), which serves as the ultimate goal for the scenario; $T$ captures the tools used in the reasoning process; and $M$ contains the metadata ($C$) of the scenario. 

By storing and retrieving experiences in this structured format, our agent learns not just to perform a task but to understand the context, common failure patterns, and successful reasoning paths. This process enables it to proactively leverage its toolkit for introspection, refining its plans by retrieving and applying relevant past experience to new, unforeseen situations.

\subsection{Experience-Driven Tool Interaction and Introspection}
\subsubsection{Vision Toolkit}
To augment the VLM's perceptual capabilities beyond its intrinsic understanding, MTRDrive is equipped with specialized Vision Toolkits. This suite of external tools allows the agent to actively probe the visual environment for specific, high-fidelity information. The primary tools include:

\textbf{Object Detection}: This tool performs standard object detection to identify and locate common traffic agents such as cars, pedestrians, and cyclists within a defined spatial range. It provides the foundational awareness of the immediate driving context.

\textbf{Open-World Vocabulary Detection}: Going beyond pre-defined categories, this tool enables the agent to search for any object or concept described by an open-vocabulary text query. This is crucial for interpreting unique or complex scenarios, such as finding "a 'No Right Turn' sign near the construction area", as shown in Figure \ref{fig:architecture}.

\textbf{Image cropping}: This tool defines a function to crop a rectangular region from an image. It requires the input image path, an output path for the new file, and the coordinates of the crop region. This function serves as the agent's “zoom-in” mechanism for meticulous examination of critical areas.

\subsubsection{Experience-driven approach to tool interaction}
While vision toolkits provide powerful perceptual abilities, a critical challenge highlighted by recent studies\cite{hu2024visualsketch, liu2025infiMMR} is that Multimodal Large Language Models (MLLMs) can develop a bias towards textual information during reasoning, leading to inappropriate or misguided tool selections. To address this, MTRDrive introduces a core innovation: an experience-driven approach that guides tool interaction. By retrieving the top-K most similar past scenarios, the model uses the documented reasoning processes ($P$) and tool usage patterns ($T$) as a strong contextual prior. This guides the VLM to make more informed and visually-grounded decisions about which tool to deploy, ensuring its actions are directly relevant to the current visual reality rather than being driven by textual artifacts. This principle of grounding tool selection in experience can be formally expressed as modeling a policy:

$$T_{\text{select}} \sim P(T \mid \mathcal{O}_{\text{current}}, \mathcal{E}_{\text{retrieved}})$$

where $T_{\text{select}}$ represents the tool selection action, $\mathcal{O}_{\text{current}}$ is the current visual observation, and $\mathcal{E}_{\text{retrieved}}$ is the context provided by the set of top-K retrieved experiences. This formulation highlights that the tool-use decision is critically informed by relevant past knowledge, not just the immediate sensory input.

\begin{figure}[htbp!]
    \centering
    \includegraphics[width=0.95\linewidth]{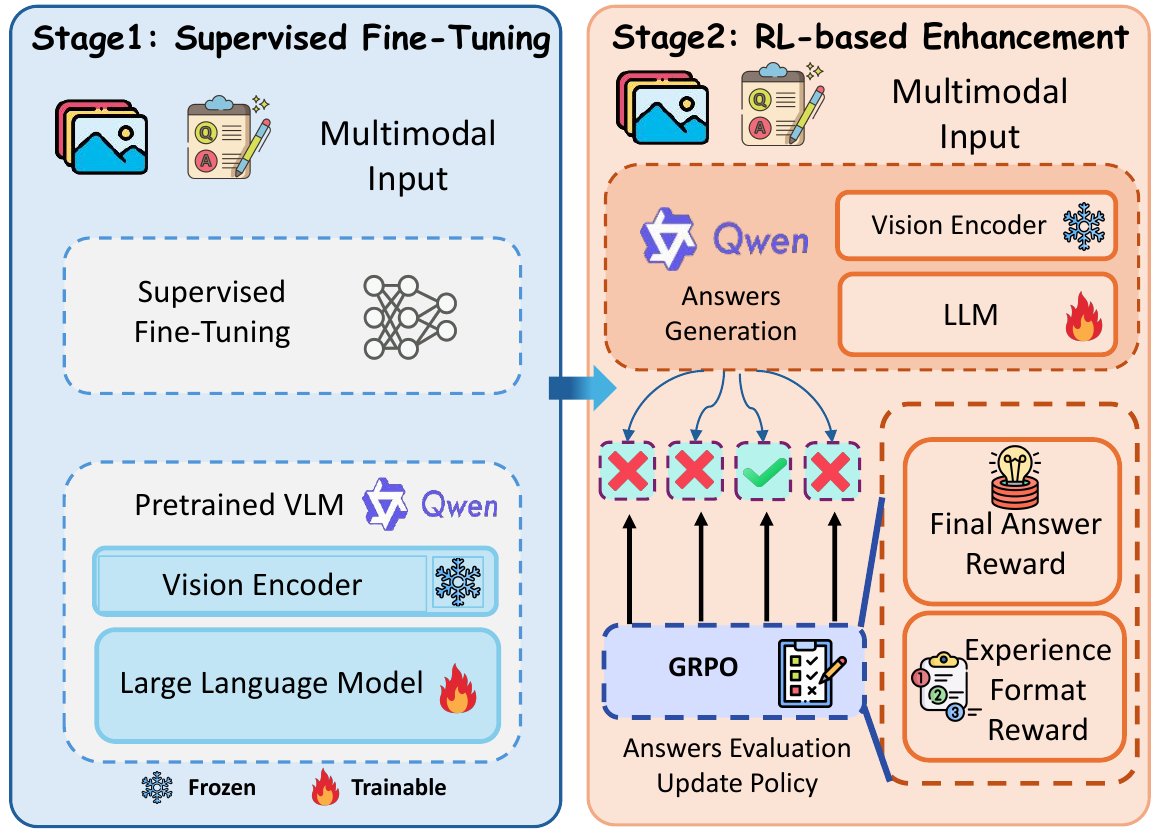}
    \caption{Two-stage Training Pipeline for MTRDrive.
This figure illustrates the two-stage training pipeline, which consists of \textbf{SFT as a Warm-up} and \textbf{RLFT for Policy Refinement}, designed to effectively train MTRDrive, a model aimed at autonomous driving tasks.}
    \label{fig:training}
\end{figure}

\subsubsection{Two-stage Training Pipeline}
To effectively train MTRDrive, we adopt a two-stage training pipeline designed to first impart foundational skills and then refine them through goal-oriented optimization.

\textbf{SFT as a Warm-up.}
The initial stage is SFT, serving as a critical warm-up phase to address the cold start problem. By fine-tuning the model, we teach it the fundamental syntax for tool use and memory integration. This process yields a competent initial policy, $\pi_{\text{ref}}$, which provides a robust starting point for the subsequent reinforcement learning phase.

\textbf{RLFT for Policy Refinement}
In the second stage, we further enhance the model's decision-making abilities using RLFT. We employ the GRPO algorithm \cite{shao2024deepseekmath} to optimize the initial policy $\pi_\theta$ based on our task-specific reward signals. GRPO evaluates multiple candidate outputs sampled from the reference policy $\pi_{ref}$ to guide the optimization. The objective is formulated as:
\begin{align}
    \mathcal{J}_\text{GRPO}(\theta) &= \mathbb{E}_{q, \{o_i\} \sim \pi_{\text{ref}}} \left[ \frac{1}{G} \sum_{i=1}^G \mathcal{J}_i - \beta \mathbb{D}_{KL}(\pi_{\theta} || \pi_{ref}) \right] \\
    \mathcal{J}_i &= \min \big( w_i A_i, \text{clip}\left(w_i, 1 - \epsilon, 1 + \epsilon\right) A_i \big)
\end{align}
where $w_i$ is the importance sampling ratio and $A_i$ is the calculated advantage.

To guide the model in learning to strategically invoke its experience, we introduce a specialized format reward function. Specifically, the format reward function $R_{\text{format}}$ is defined as:
\begin{equation}
R_{\text{format}} =
\begin{cases}
    1.0, & \text{if experience is correctly used} \\
    0.5, & \text{if experience is correctly omitted} \\
    0, & \text{otherwise}
\end{cases}
\end{equation}
The correct choice is determined based on a heuristic related to the relevance of retrieved experiences for the current task.

The complete reward function is a combination of the format reward and a task accomplishment reward, defined as:
\begin{equation}
R = \lambda \cdot R_{\text{format}} + R_{\text{Acc}}
\end{equation}
where $R_{\text{Acc}}$ represents the accuracy reward for the final driving plan, and $\lambda$ is a hyperparameter used to adjust the balance between the two reward components. Essentially, the reward function we designed provides the model with an explicit signal for the meta-cognitive task of deciding when to seek external knowledge. This enables the model to actively reflect on the utility of its experience during the reasoning process, similar to how humans learn to draw upon past experiences when facing novel or challenging situations.





\begin{table*}[htbp!]
\centering
\caption{Performance Comparison on NAVSIM (SFT) and ROADWork (Zero-Shot) Benchmarks}
\label{tab:unified_performance}
\begin{tabular}{>{\centering\arraybackslash}p{2cm}l ccc ccc}
\toprule
\multirow{2}{*}{\textbf{Dataset}} & \multirow{2}{*}{\textbf{Vision Language Models}} & \multicolumn{3}{c}{\textbf{Driving Metrics (\%) $\uparrow$}} & \multicolumn{3}{c}{\textbf{Planning (\%) $\uparrow$}} \\
\cmidrule(lr){3-5} \cmidrule(lr){6-8}
& & \textbf{Risk Assess.} & \textbf{Reason.} & \textbf{Scene Aware.} & \textbf{Path.} & \textbf{Speed.} & \textbf{Acc.} \\
\midrule
\multirow{6}{*}{\textbf{\makecell{NAVSIM \\(SFT)}}} & Qwen2.5-VL-72B & 82.9 & 81.1 & 84.0 & 70.4 & 48.2 & 37.8 \\
& Qwen2.5-VL-32B & \textbf{85.7} & \textbf{85.3} & \textbf{86.4} & 67.1 & 48.2 & 35.9 \\
& InternVL3-8B & 81.2 & 79.7 & 81.6 & 41.8 & 34.8 & 16.5 \\
& Qwen2.5-VL-7B & 77.0 & 74.9 & 76.6 & 15.70 & 41.6 & 8.1 \\
& Qwen2.5-VL-3B & 74.1 & 72.1 & 73.5 & 4.4 & 30.7 & 1.6 \\
\cline{2-8}
\rowcolor{gray!20}
& MTRDrive (Ours) & 79.8 & 78.8 & 80.8 & \textbf{93.1} & \textbf{84.6} & \textbf{82.6} \\
\midrule
\multirow{6}{*}{\textbf{\makecell{ROADWork \\ (Zero-shot)}}} & Qwen2.5-VL-72B & 83.6 & 80.9 & 84.8 & \textbf{47.6} & 61.1 & 29.7 \\
& Qwen2.5-VL-32B & \textbf{85.7} & \textbf{85.3} & \textbf{86.4} & 47.1 & 45.5 & 23.5 \\
& InternVL3-8B & 77.9 & 73.7 & 77.7 & 33.5 & 63.9 & 23.1 \\
& Qwen2.5-VL-7B & 71.8 & 66.6 & 70.5 & 15.23 & 63.8 & 9.2 \\
& Qwen2.5-VL-3B & 64.5 & 64.5 & 61.7 & 6.1 & 29.5 & 0.5 \\
\cline{2-8}
\rowcolor{gray!20}
& MTRDrive (Ours) & 80.2 & 79.6 & 80.3 & 44.2 & \textbf{72.1} & \textbf{33.5} \\
\bottomrule
\end{tabular}
\end{table*}

\begin{figure*}[t!]
    \centering
    \includegraphics[width=1.0\linewidth]{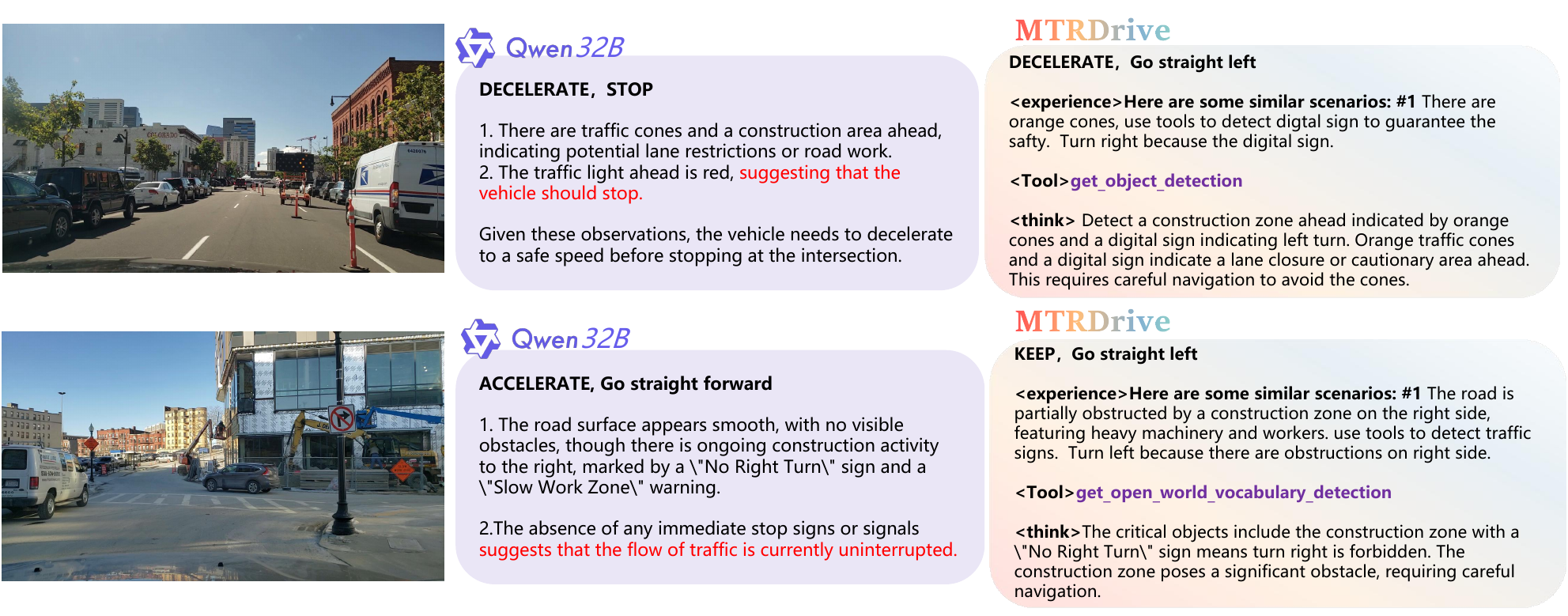}
    \caption{Qualitative comparison of MTRDrive versus the Qwen-32B baseline in challenging real-world scenarios. The examples highlight how MTRDrive's interactive reasoning, which leverages past experiences and tools, mitigates common VLM failures like visual hallucination (top) and flawed reasoning (bottom), leading to safer and more accurate driving plans.}
    \label{fig:visualize}
\end{figure*}

\section{Experiments}

In this section, we conduct a series of extensive experiments to validate the effectiveness of our \textbf{MTRDrive} framework. Our experimental design is centered around answering the following core questions to demonstrate the efficacy and robustness of our approach:

\begin{enumerate}
\item \textbf{Q1: Can our interactive reasoning approach significantly improve model generalization and effectively mitigate hallucinations?} 
\item \textbf{Q2: Can MTRDrive demonstrate strong zero-shot generalization on our newly constructed RoadWork-VLM benchmark?} 

\item \textbf{Q3: How does our two-stage training strategy enable the model to learn efficient and correct interactive behaviors?} 

\end{enumerate}

\begin{table*}[htbp!]
\centering
\caption{Ablation Study on NAVSIM (SFT) and ROADWork (Zero-Shot) Benchmarks}
\label{tab:high-level-ablation}
\begin{tabular}{>{\centering\arraybackslash}p{2.3cm} l ccc ccc ccc}
\toprule
\multirow{2}{*}{\textbf{Dataset}} & \multirow{2}{*}{\textbf{Model Variant}} & 
\multirow{2}{*}{\textbf{SFT}} & \multirow{2}{*}{\makecell{\textbf{Acc.}\\\textbf{Reward}}} & \multirow{2}{*}{\makecell{\textbf{Driving}\\\textbf{Exp.}}} &
\multicolumn{3}{c}{\textbf{Driving Metrics (\%) $\uparrow$}} & \multicolumn{3}{c}{\textbf{Planning (\%) $\uparrow$}} \\
\cmidrule(lr){6-8} \cmidrule(lr){9-11}
& & & & & \textbf{Risk Assess.} & \textbf{Reason.} & \textbf{Scene Aware.} & \textbf{Path.} & \textbf{Speed.} & \textbf{Acc.} \\
\midrule

\multirow{4}{*}{\textbf{\shortstack{NAVSIM \\ (SFT)}}} 
& BaseModel       & \ding{55} & \ding{55} & \ding{55} & 74.1 & 72.1 & 73.5 & 4.4  & 30.7 & 1.6 \\
& +SFT            & \ding{51} & \ding{55} & \ding{55} & 78.2 & 78.1 & 78.7 & 92.8 & 83.4 & 79.9 \\
& +SFT+GRPO       & \ding{51} & \ding{51} & \ding{55} & 79.9 & 78.8 & 80.6 & 92.0 & 84.2 & 79.6 \\
\cline{2-11}
\rowcolor{gray!20}
& MTRDrive (Ours) & \ding{51} & \ding{51} & \ding{51} & \textbf{79.8} & \textbf{78.8} & \textbf{80.8} & \textbf{93.1} & \textbf{84.6} & \textbf{82.6} \\
\midrule

\multirow{4}{*}{\textbf{\shortstack{ROADWork \\ (Zero-Shot)}}} 
& BaseModel       & \ding{55} & \ding{55} & \ding{55} & 64.5 & 64.5 & 61.7& 6.1  & 29.5 & 0.5 \\
& +SFT(NAVSIM)            & \ding{51} & \ding{55} & \ding{55} & 75.0 & 75.9 & 75.4 & 48.6 & 27.5 & 13.3 \\
& +SFT+GRPO (NAVSIM)     & \ding{51} & \ding{51} & \ding{55} & 79.2 & 79.4 & 80.4 & 17.3 & 36.4 & 17.3 \\
\cline{2-11}
\rowcolor{gray!20}
& MTRDrive (Ours) & \ding{51} & \ding{51} & \ding{51} & \textbf{80.2} & \textbf{79.6} & \textbf{80.3} & 44.2 & \textbf{72.1} & \textbf{33.5} \\
\bottomrule
\end{tabular}
\end{table*}

Beyond these core questions, we conduct a supplementary experiment on the NAVSIM dataset to further investigate the impact of high-level decisions on closed-loop trajectory prediction. This study highlights the crucial role of high-level planning in generating accurate and reliable trajectories.

\subsection{Setup}
\subsubsection{Datasets}
We conduct our experiments on two distinct benchmarks: NAVSIM\cite{dauner2024navsim} and ROADWork\cite{ghosh2024roadwork}. NAVSIM\cite{dauner2024navsim} is a large-scale real-world autonomous driving dataset designed for non-reactive simulation and benchmarking. Built upon OpenScene, it specifically focuses on challenging scenarios involving dynamic intention changes, while filtering out trivial cases such as stationary scenes or constant-speed driving. To align with high-level decision making, we further process the NAVSIM trajectories to compute the vehicle's dynamic characteristics—such as speed and corresponding angular changes, and finally derive high-level speed and path plans.

However, due to the relative simplicity of its scenarios, NAVSIM alone is insufficient to validate an agent's ability to handle the complexities of real-world driving. To address this, we specifically test our model's generalization to more intricate, unseen scenarios like construction zones.

To rigorously evaluate our agent's zero-shot generalization and reasoning capabilities in these complex environments, we introduce Roadwork-VLM, a new benchmark meticulously processed from the original ROADWork dataset\cite{ghosh2024roadwork}. We leveraged the powerful Qwen2.5-VL-72B\cite{bai2025qwen25} model to re-annotate the entire dataset, generating detailed scenario descriptions, high-level navigation instructions, and full Chain-of-Thought reasoning sequences based on the provided image and trajectory information. This process results in a complete, human-like VLM dataset for end-to-end driving, enabling us to test an agent's ability to perform complex, high-level behavioral decisions in a zero-shot setting. We will open-source Roadwork-VLM to foster further research in this critical area.



\subsubsection{Metrics}
We evaluate our framework's performance on two distinct aspects of autonomous driving: long-term high-level planning and short-term trajectory prediction. 

For high-level planning and reasoning, we use a combination of metrics. Our primary metric is High-Level Planning Accuracy, which strictly assesses the correctness of an agent's strategic decision. A plan is only considered correct if the entire meta-action—encompassing both longitudinal speed and lateral path—exactly matches the ground truth. To analyze the thought process behind these decisions, we adopt the methodology from DriveLMM-o1 \cite{ishaq2025drivelmmo1}, where GPT-4o-mini serves as an automated judge to evaluate the model's textual output on \textbf{Risk Assessment}, \textbf{Commonsense Reasoning}, and \textbf{Scene Awareness}.

For the fine-grained task of short-term trajectory prediction, we use the standard metric from the NAVSIM benchmark. The Predictive Driver Model Score (PDMS) is employed to measure the point-wise accuracy and physical realism of the agent's predicted closed-loop trajectory.

\subsubsection{Implementation Details}
We build our MTRDrive agent upon the Qwen2.5-VL-3B\cite{bai2025qwen25}, an open-source Vision-Language Model that provides a strong balance of performance and efficiency. All experiments are conducted on 16 NVIDIA H20 GPUs.

Our two-stage training pipeline is configured as follows. For the initial SFT stage, we fine-tune the model for 2 epochs using the AdamW optimizer with a learning rate of $4e-5$ and a global batch size of 2. During this stage, the parameters of the vision encoder are kept frozen to preserve its pre-trained features. In the subsequent RLFT stage, we employ the GRPO algorithm with a group size of 8 and a KL penalty coefficient $\beta$ of 0.02. All stages utilize a cosine annealing learning rate scheduler to ensure stable convergence.

\subsection{Experiment Results on High level planning}
\subsubsection{Main Results}
As presented in Table \ref{tab:unified_performance}, our results highlight MTRDrive's decisive advantage in high-level planning. On the NAVSIM (SFT) benchmark, our model's $82.6\%$ accuracy more than doubles the $37.8\%$ of Qwen2.5-VL-72B. This superiority is maintained in the challenging zero-shot ROADWork setting, where MTRDrive achieves $33.5\%$ accuracy, maintaining a clear lead over the strongest baseline ($29.7\%$).

This success is particularly noteworthy as it does not simply stem from superior general reasoning. While massive VLMs exhibit high raw reasoning scores, our model is exceptionally effective at translating its understanding into correct, actionable plans. Crucially, in the zero-shot setting, MTRDrive's reasoning scores become highly competitive. This suggests our experience-retrieval mechanism is a key factor for robustly making sound decisions in novel environments, validating our core design principle.


Qualitatively, as illustrated in Fig.\ref{fig:visualize}, MTRDrive successfully navigates challenging, real-world corner cases where the base VLM fails. In these complex scenarios, the Qwen-32B model often struggles with flawed reasoning, such as hallucinating critical objects or failing to grasp the safety implications of the scene, which leads to incorrect and potentially dangerous driving plans. In contrast, MTRDrive adeptly retrieves relevant past experiences and invokes its tools to acquire crucial decision-making information. By grounding its reasoning in this retrieved context, it avoids hallucinations and correctly interprets complex situations, such as navigating a construction zone or obeying a prohibitive traffic sign.

\begin{table}[htbp!]
\centering
\footnotesize
\caption{
 NAVSIM \texttt{navtest} closed-loop results.
 \textbf{Bold} = best overall; $\dagger$ = fine-tuned on NAVSIM labels.
}
\label{tab:navsim}
\setlength{\tabcolsep}{4pt}
\renewcommand{\arraystretch}{1.05}
\begin{tabular}{
 >{\raggedright\arraybackslash}p{2.5cm}
 S[table-format=2.1]
 S[table-format=2.1]
 S[table-format=2.1]
 S[table-format=2.1]
 S[table-format=2.1]
 S[table-format=2.1]
}
\toprule
\textbf{Method} & {\textbf{NC}} & {\textbf{DAC}} & {\textbf{TTC}} & {\textbf{CF}} & {\textbf{EP}} & {\textbf{PDMS}} \\
\midrule
VADv2 \cite{vad} & 97.2 & 89.1 & 91.6 & 100.0 & 76.0 & 80.9 \\
DrivingGPT \cite{drivinggpt} & 98.9 & 90.7 & 94.9 & 95.6 & 79.7 & 82.4 \\
UniAD \cite{uniad} & 97.8 & 91.9 & 92.9 & 100.0 & 78.8 & 83.4 \\
TransFuser \cite{transfuser} & 97.7 & 92.8 & 92.8 & 100.0 & 79.2 & 84.0 \\
DiffusionDrive \cite{diffusiondrive} & 98.2 & 96.2 & 94.7 & 100.0 & 82.2 & 88.1 \\
WoTE \cite{wote} & 98.5 & 96.8 & 94.9 & 99.9 & 81.9 & 88.3 \\
\midrule
\multicolumn{7}{c}{\textit{VLM-based methods}} \\
\midrule
InternVL3-2B$^\dagger$ & 97.2 & 87.4 & 91.8 & 100.0 & 78.9 & 78.9 \\
QwenVL2.5-3B$^\dagger$ & 97.7 & 93.1 & 92.3 & 100.0 & 73.0 & 84.2 \\
\mbox{+High-level Decisions} & 98.2 & 94.6 & 94.3 & 100.0 & 80.6 & 86.3 \\
+GRPO & 96.8 & 95.4 & 90.5 & 100.0 & 86.1 & 87.3 \\
\rowcolor{gray!15}
\textbf{MTRDrive} & 97.3 & \textbf{95.8} & 91.2 & 100.0 & \textbf{86.8} & \textbf{88.3} \\
\bottomrule
\end{tabular}
\end{table}
\subsubsection{Ablation Study}
To dissect the contribution of each component in our framework, we conducted a thorough ablation study, with results presented in Table \ref{tab:high-level-ablation}. We analyze the incremental impact of Supervised Fine-Tuning (SFT), Reinforcement Learning (GRPO) with our proposed reward, and the Driving Experience module.

The study first reveals that the base VLM, without any training, fails at the planning task ($1.6\%$ accuracy on NAVSIM). The introduction of SFT provides the most significant initial performance boost, increasing accuracy to $79.9\%$ on NAVSIM by teaching the model the fundamental syntax and structure of the planning task. The subsequent addition of GRPO further refines the policy, leading to improved reasoning scores and a notable increase in zero-shot planning accuracy on ROADWork from 13.3\% to 17.3\%.

Most critically, the final addition of the Driving Experience module demonstrates its indispensable role in generalization. While providing a solid boost on the in-domain NAVSIM dataset to our final 82.6\% accuracy, its impact on the zero-shot ROADWork benchmark is transformative. It nearly doubles the planning accuracy from 17.3\% to our final 33.5\%. This result unequivocally proves that the experience-retrieval mechanism is the key component that empowers the model to apply its learned skills effectively to new and unseen driving scenarios.


\subsection{Impact of High-Level Planning on Trajectory Prediction}
Beyond evaluating high-level planning itself, we conduct a supplementary experiment on the NAVSIM dataset to investigate the impact of these high-level decisions on closed-loop trajectory prediction. This study, detailed in Table \ref{tab:navsim}, highlights the crucial role of structured reasoning in generating accurate and reliable trajectories.

Our analysis begins with a strong VLM baseline, QwenVL2.5-3B. This model is fine-tuned to function as a pure trajectory predictor, directly outputting waypoints without any Chain-of-Thought reasoning, and achieves a respectable PDMS of 84.2. However, the key insight emerges when we introduce coarse-to-fine strategy. By prompting the model to first predict a high-level meta-action before generating the trajectory, the performance significantly improves to a PDMS of 86.3. This single modification, focusing purely on hierarchical planning, is sufficient to outperform highly specialized methods like UniAD (83.4) and TransFuser (84.0). This result strongly validates our hypothesis that decomposing the planning task and leveraging high-level reasoning is critical for improving low-level trajectory quality.

Building upon this foundation, the subsequent integration of our GRPO training strategy and the experience-retrieval mechanism further enhances performance. Our full MTRDrive model achieves a final PDMS of 88.3. This result not only demonstrates the synergistic benefits of our complete framework but also matches the state-of-the-art performance of specialized methods like WoTE on this benchmark. While our method achieves top-tier performance through superior high-level reasoning, we believe further gains are possible by enhancing the trajectory decoder like Recogdrive\cite{li2025recogdrive} and AutoVLA\cite{zhou2025autovla}. Exploring a more sophisticated decoder architecture to improve the fine-grained quality of the trajectory output remains a promising direction for our future research.

\section{CONCLUSIONS}

We have presented MTRDrive, an interactive reasoning framework for end-to-end autonomous driving that successfully mitigates hallucinations and improves out-of-distribution generalization in Vision-Language Models. By integrating memory-based experience retrieval with a dynamic toolkit, MTRDrive enables grounded, proactive decision-making through a closed-loop, human-like reasoning process. To validate its generalization capabilities, we also introduced a new benchmark focused on complex Roadwork construction scenarios.

Ultimately, MTRDrive advances autonomous driving systems from passive perception to active, experience-guided reasoning, moving closer to reliable and interpretable deployment. This work lays a solid foundation for future VLM-based agents capable of handling complex, long-tail driving scenarios. 

For future work, we plan to explore the integration of different specialized decoders with our framework. This will allow for a deeper investigation into how the synergy between memory retrieval and tool use can be optimized to enhance the fine-grained quality of VLM-based trajectory planning.

\bibliographystyle{IEEEtran}
\bibliography{ref}

\begin{thebibliography}{10}
\providecommand{\url}[1]{#1}
\csname url@samestyle\endcsname
\providecommand{\newblock}{\relax}
\providecommand{\bibinfo}[2]{#2}
\providecommand{\BIBentrySTDinterwordspacing}{\spaceskip=0pt\relax}
\providecommand{\BIBentryALTinterwordstretchfactor}{4}
\providecommand{\BIBentryALTinterwordspacing}{\spaceskip=\fontdimen2\font plus
\BIBentryALTinterwordstretchfactor\fontdimen3\font minus \fontdimen4\font\relax}
\providecommand{\BIBforeignlanguage}[2]{{%
\expandafter\ifx\csname l@#1\endcsname\relax
\typeout{** WARNING: IEEEtran.bst: No hyphenation pattern has been}%
\typeout{** loaded for the language `#1'. Using the pattern for}%
\typeout{** the default language instead.}%
\else
\language=\csname l@#1\endcsname
\fi
#2}}
\providecommand{\BIBdecl}{\relax}
\BIBdecl

\bibitem{jiang2025vlasurvey}
S.~Jiang, Z.~Huang, K.~Qian, Z.~Luo, T.~Zhu, Y.~Zhong, Y.~Tang, M.~Kong, Y.~Wang, S.~Jiao \emph{et~al.}, ``A survey on vision-language-action models for autonomous driving,'' \emph{arXiv preprint arXiv:2506.24044}, 2025.

\bibitem{shi2024effocc}
Y.~Shi, K.~Jiang, K.~Wang, K.~Qian, Y.~Wang, J.~Li, T.~Wen, M.~Yang, Y.~Xu, and D.~Yang, ``Effocc: A minimal baseline for efficient fusion-based 3d occupancy network,'' \emph{arXiv e-prints}, pp. arXiv--2406, 2024.

\bibitem{qian2024priormotion}
K.~Qian, J.~Miao, X.~Jiao, Z.~Luo, Z.~Fu, Y.~Shi, Y.~Wang, K.~Jiang, and D.~Yang, ``Priormotion: Generative class-agnostic motion prediction with raster-vector motion field priors,'' \emph{arXiv preprint arXiv:2412.04020}, 2024.

\bibitem{qian2025lego}
K.~Qian, J.~Miao, Z.~Luo, Z.~Fu, Y.~Shi, Y.~Wang, K.~Jiang, M.~Yang, D.~Yang \emph{et~al.}, ``Lego-motion: Learning-enhanced grids with occupancy instance modeling for class-agnostic motion prediction,'' \emph{arXiv preprint arXiv:2503.07367}, 2025.

\bibitem{qian2024spider}
Z.~Qian, K.~Jiang, Z.~Cao, K.~Qian, Y.~Xu, W.~Zhou, and D.~Yang, ``Spider: Self-driving planners and intelligent decision-making engines with reusability,'' in \emph{2024 IEEE 27th International Conference on Intelligent Transportation Systems (ITSC)}.\hskip 1em plus 0.5em minus 0.4em\relax IEEE, 2024, pp. 937--944.

\bibitem{fu2025enhancing}
Z.~Fu, H.~Lin, K.~Qian, T.~Wen, H.~Gao, Z.~Zhong, and D.~Yang, ``Enhancing autonomous vehicle planning with a robust fault-tolerant mechanism for action-induced agent detection,'' in \emph{ICASSP 2025-2025 IEEE International Conference on Acoustics, Speech and Signal Processing (ICASSP)}.\hskip 1em plus 0.5em minus 0.4em\relax IEEE, 2025, pp. 1--5.

\bibitem{qian2025agentthink}
K.~Qian, S.~Jiang, Y.~Zhong, Z.~Luo, Z.~Huang, T.~Zhu, K.~Jiang, M.~Yang, Z.~Fu, J.~Miao \emph{et~al.}, ``Agentthink: A unified framework for tool-augmented chain-of-thought reasoning in vision-language models for autonomous driving,'' \emph{arXiv preprint arXiv:2505.15298}, 2025.

\bibitem{nie2024reason2drive}
M.~Nie, R.~Peng, C.~Wang, X.~Cai, J.~Han, H.~Xu, and L.~Zhang, ``Reason2drive: Towards interpretable and chain-based reasoning for autonomous driving,'' in \emph{European Conference on Computer Vision}.\hskip 1em plus 0.5em minus 0.4em\relax Springer, 2024, pp. 292--308.

\bibitem{shao2024lmdrive}
H.~Shao, Y.~Hu, L.~Wang, G.~Song, S.~L. Waslander, Y.~Liu, and H.~Li, ``Lmdrive: Closed-loop end-to-end driving with large language models,'' in \emph{Proceedings of the IEEE/CVF Conference on Computer Vision and Pattern Recognition}, 2024, pp. 15\,120--15\,130.

\bibitem{sima2024drivelm}
C.~Sima, K.~Renz, K.~Chitta, L.~Chen, H.~Zhang, C.~Xie, J.~Bei{\ss}wenger, P.~Luo, A.~Geiger, and H.~Li, ``Drivelm: Driving with graph visual question answering,'' in \emph{European conference on computer vision}.\hskip 1em plus 0.5em minus 0.4em\relax Springer, 2024, pp. 256--274.

\bibitem{li2025recogdrive}
Y.~Li, K.~Xiong, X.~Guo, F.~Li, S.~Yan, G.~Xu, L.~Zhou, L.~Chen, H.~Sun, B.~Wang \emph{et~al.}, ``Recogdrive: A reinforced cognitive framework for end-to-end autonomous driving,'' \emph{arXiv preprint arXiv:2506.08052}, 2025.

\bibitem{fu2025orion}
H.~Fu, D.~Zhang, Z.~Zhao, J.~Cui, D.~Liang, C.~Zhang, D.~Zhang, H.~Xie, B.~Wang, and X.~Bai, ``Orion: A holistic end-to-end autonomous driving framework by vision-language instructed action generation,'' \emph{arXiv preprint arXiv:2503.19755}, 2025.

\bibitem{zhou2025autovla}
Z.~Zhou, T.~Cai, S.~Z. Zhao, Y.~Zhang, Z.~Huang, B.~Zhou, and J.~Ma, ``Autovla: A vision-language-action model for end-to-end autonomous driving with adaptive reasoning and reinforcement fine-tuning,'' \emph{arXiv preprint arXiv:2506.13757}, 2025.

\bibitem{ghosh2024roadwork}
A.~Ghosh, S.~Zheng, R.~Tamburo, K.~Vuong, J.~Alvarez-Padilla, H.~Zhu, M.~Cardei, N.~Dunn, C.~Mertz, and S.~G. Narasimhan, ``Roadwork dataset: learning to recognize, observe, analyze and drive through work zones,'' \emph{arXiv preprint arXiv:2406.07661}, 2024.

\bibitem{qian2024fasionad}
K.~Qian, Z.~Ma, Y.~He, Z.~Luo, T.~Shi, T.~Zhu, J.~Li, J.~Wang, Z.~Chen, X.~He \emph{et~al.}, ``Fasionad: Fast and slow fusion thinking systems for human-like autonomous driving with adaptive feedback,'' \emph{arXiv preprint arXiv:2411.18013}, 2024.

\bibitem{qian2025fasionad++}
K.~Qian, Z.~Luo, S.~Jiang, Z.~Huang, J.~Miao, Z.~Ma, T.~Zhu, J.~Li, Y.~He, Z.~Fu \emph{et~al.}, ``Fasionad++: Integrating high-level instruction and information bottleneck in fat-slow fusion systems for enhanced safety in autonomous driving with adaptive feedback,'' \emph{arXiv preprint arXiv:2503.08162}, 2025.

\bibitem{mao2023gptdriver}
J.~Mao, Y.~Qian, J.~Ye, H.~Zhao, and Y.~Wang, ``Gpt-driver: Learning to drive with gpt,'' \emph{arXiv preprint arXiv:2310.01415}, 2023.

\bibitem{jiang2024senna}
B.~Jiang, S.~Chen, B.~Liao, X.~Zhang, W.~Yin, Q.~Zhang, C.~Huang, W.~Liu, and X.~Wang, ``Senna: Bridging large vision-language models and end-to-end autonomous driving,'' \emph{arXiv preprint arXiv:2410.22313}, 2024.

\bibitem{xu2024drivegpt4}
Z.~Xu, Y.~Zhang, E.~Xie, Z.~Zhao, Y.~Guo, K.-Y.~K. Wong, Z.~Li, and H.~Zhao, ``Drivegpt4: Interpretable end-to-end autonomous driving via large language model,'' \emph{IEEE Robotics and Automation Letters}, 2024.

\bibitem{zhou2025opendrivevla}
X.~Zhou, X.~Han, F.~Yang, Y.~Ma, and A.~C. Knoll, ``Opendrivevla: Towards end-to-end autonomous driving with large vision language action model,'' \emph{arXiv preprint arXiv:2503.23463}, 2025.

\bibitem{hwang2024emma}
J.-J. Hwang, R.~Xu, H.~Lin, W.-C. Hung, J.~Ji, K.~Choi, D.~Huang, T.~He, P.~Covington, B.~Sapp \emph{et~al.}, ``Emma: End-to-end multimodal model for autonomous driving,'' \emph{arXiv preprint arXiv:2410.23262}, 2024.

\bibitem{renz2024carllava}
K.~Renz, L.~Chen, A.-M. Marcu, J.~H{\"u}nermann, B.~Hanotte, A.~Karnsund, J.~Shotton, E.~Arani, and O.~Sinavski, ``Carllava: Vision language models for camera-only closed-loop driving,'' \emph{arXiv preprint arXiv:2406.10165}, 2024.

\bibitem{renz2025simlingo}
K.~Renz, L.~Chen, E.~Arani, and O.~Sinavski, ``Simlingo: Vision-only closed-loop autonomous driving with language-action alignment,'' in \emph{Proceedings of the Computer Vision and Pattern Recognition Conference}, 2025, pp. 11\,993--12\,003.

\bibitem{wang2024drivecot}
T.~Wang, E.~Xie, R.~Chu, Z.~Li, and P.~Luo, ``Drivecot: Integrating chain-of-thought reasoning with end-to-end driving,'' \emph{arXiv preprint arXiv:2403.16996}, 2024.

\bibitem{mao2023agentdriver}
J.~Mao, J.~Ye, Y.~Qian, M.~Pavone, and Y.~Wang, ``A language agent for autonomous driving,'' \emph{arXiv preprint arXiv:2311.10813}, 2023.

\bibitem{zhao2025sce2drivex}
R.~Zhao, Q.~Yuan, J.~Li, H.~Hu, Y.~Li, C.~Zheng, and F.~Gao, ``Sce2drivex: A generalized mllm framework for scene-to-drive learning,'' \emph{arXiv preprint arXiv:2502.14917}, 2025.

\bibitem{chi2025impromptu}
H.~Chi, H.-a. Gao, Z.~Liu, J.~Liu, C.~Liu, J.~Li, K.~Yang, Y.~Yu, Z.~Wang, W.~Li \emph{et~al.}, ``Impromptu vla: Open weights and open data for driving vision-language-action models,'' \emph{arXiv preprint arXiv:2505.23757}, 2025.

\bibitem{dauner2024navsim}
D.~Dauner, M.~Hallgarten, T.~Li, X.~Weng, Z.~Huang, Z.~Yang, H.~Li, I.~Gilitschenski, B.~Ivanovic, M.~Pavone \emph{et~al.}, ``Navsim: Data-driven non-reactive autonomous vehicle simulation and benchmarking,'' \emph{Advances in Neural Information Processing Systems}, vol.~37, pp. 28\,706--28\,719, 2024.

\bibitem{jia2024bench2drive}
X.~Jia, Z.~Yang, Q.~Li, Z.~Zhang, and J.~Yan, ``Bench2drive: Towards multi-ability benchmarking of closed-loop end-to-end autonomous driving,'' \emph{Advances in Neural Information Processing Systems}, vol.~37, pp. 819--844, 2024.

\bibitem{wang2023drivemlm}
W.~Wang, J.~Xie, C.~Hu, H.~Zou, J.~Fan, W.~Tong, Y.~Wen, S.~Wu, H.~Deng, Z.~Li \emph{et~al.}, ``Drivemlm: Aligning multi-modal large language models with behavioral planning states for autonomous driving,'' \emph{arXiv preprint arXiv:2312.09245}, 2023.

\bibitem{bai2025qwen25}
S.~Bai, K.~Chen, X.~Liu, J.~Wang, W.~Ge, S.~Song, K.~Dang, P.~Wang, S.~Wang, J.~Tang \emph{et~al.}, ``Qwen2. 5-vl technical report,'' \emph{arXiv preprint arXiv:2502.13923}, 2025.

\bibitem{radford2021clip}
A.~Radford, J.~W. Kim, C.~Hallacy, A.~Ramesh, G.~Goh, S.~Agarwal, G.~Sastry, A.~Askell, P.~Mishkin, J.~Clark \emph{et~al.}, ``Learning transferable visual models from natural language supervision,'' in \emph{International conference on machine learning}.\hskip 1em plus 0.5em minus 0.4em\relax PmLR, 2021, pp. 8748--8763.

\bibitem{hu2024visualsketch}
Y.~Hu, W.~Shi, X.~Fu, D.~Roth, M.~Ostendorf, L.~Zettlemoyer, N.~A. Smith, and R.~Krishna, ``Visual sketchpad: Sketching as a visual chain of thought for multimodal language models,'' \emph{Advances in Neural Information Processing Systems}, vol.~37, pp. 139\,348--139\,379, 2024.

\bibitem{liu2025infiMMR}
Z.~Liu, Y.~Liu, G.~Zhu, C.~Xie, Z.~Li, J.~Yuan, X.~Wang, Q.~Li, S.-C. Cheung, S.~Zhang \emph{et~al.}, ``Infi-mmr: Curriculum-based unlocking multimodal reasoning via phased reinforcement learning in multimodal small language models,'' \emph{arXiv preprint arXiv:2505.23091}, 2025.

\bibitem{shao2024deepseekmath}
Z.~Shao, P.~Wang, Q.~Zhu, R.~Xu, J.~Song, X.~Bi, H.~Zhang, M.~Zhang, Y.~Li, Y.~Wu \emph{et~al.}, ``Deepseekmath: Pushing the limits of mathematical reasoning in open language models,'' \emph{arXiv preprint arXiv:2402.03300}, 2024.

\bibitem{ishaq2025drivelmmo1}
A.~Ishaq, J.~Lahoud, K.~More, O.~Thawakar, R.~Thawkar, D.~Dissanayake, N.~Ahsan, Y.~Li, F.~S. Khan, H.~Cholakkal \emph{et~al.}, ``Drivelmm-o1: A step-by-step reasoning dataset and large multimodal model for driving scenario understanding,'' \emph{arXiv preprint arXiv:2503.10621}, 2025.

\bibitem{vad}
S.~Chen, B.~Jiang, H.~Gao, B.~Liao, Q.~Xu, Q.~Zhang, C.~Huang, W.~Liu, and X.~Wang, ``Vadv2: End-to-end vectorized autonomous driving via probabilistic planning,'' \emph{arXiv preprint arXiv:2402.13243}, 2024.

\bibitem{drivinggpt}
Y.~Chen, Y.~Wang, and Z.~Zhang, ``Drivinggpt: Unifying driving world modeling and planning with multi-modal autoregressive transformers,'' \emph{arXiv preprint arXiv:2412.18607}, 2024.

\bibitem{uniad}
Y.~Hu, J.~Yang, L.~Chen, K.~Li, C.~Sima, X.~Zhu, S.~Chai, S.~Du, T.~Lin, W.~Wang \emph{et~al.}, ``Planning-oriented autonomous driving,'' in \emph{Proceedings of the IEEE/CVF conference on computer vision and pattern recognition}, 2023, pp. 17\,853--17\,862.

\bibitem{transfuser}
K.~Chitta, A.~Prakash, B.~Jaeger, Z.~Yu, K.~Renz, and A.~Geiger, ``Transfuser: Imitation with transformer-based sensor fusion for autonomous driving,'' \emph{IEEE transactions on pattern analysis and machine intelligence}, vol.~45, no.~11, pp. 12\,878--12\,895, 2022.

\bibitem{diffusiondrive}
B.~Liao, S.~Chen, H.~Yin, B.~Jiang, C.~Wang, S.~Yan, X.~Zhang, X.~Li, Y.~Zhang, Q.~Zhang \emph{et~al.}, ``Diffusiondrive: Truncated diffusion model for end-to-end autonomous driving,'' in \emph{Proceedings of the Computer Vision and Pattern Recognition Conference}, 2025, pp. 12\,037--12\,047.

\bibitem{wote}
Y.~Li, Y.~Wang, Y.~Liu, J.~He, L.~Fan, and Z.~Zhang, ``End-to-end driving with online trajectory evaluation via bev world model,'' \emph{arXiv preprint arXiv:2504.01941}, 2025.

\end{thebibliography}

\end{document}